\documentclass[sigconf]{acmart}

\copyrightyear{2023}
\acmYear{2023}
\setcopyright{acmlicensed}
\acmConference[MM '23] {Proceedings of the 31st ACM International Conference on Multimedia}{October 29--November 3, 2023}{Ottawa, ON, Canada.}
\acmBooktitle{Proceedings of the 31st ACM International Conference on Multimedia (MM '23), October 29--November 3, 2023, Ottawa, ON, Canada}
\acmPrice{15.00}
\acmISBN{979-8-4007-0108-5/23/10}
\acmDOI{10.1145/3581783.3612502}

\usepackage{multirow}
\usepackage{colortbl}
\usepackage{balance}

\AtBeginDocument{%
  }

\begin{document}

\title{Induction Network: Audio-Visual Modality Gap-Bridging for Self-Supervised Sound Source Localization}

\author{Tianyu Liu}
\orcid{0000-0002-9098-5944}
\affiliation{
  \institution{Northwestern Polytechnical University}
  \city{Xi'an}
  \state{Shanxi}
  \country{China}
}
\affiliation{
  \institution{Ningbo Institute of Northwestern Polytechnical University}
  \city{Ningbo}
  \state{Zhejiang}
  \country{China}
}
\email{reallty@mail.nwpu.edu.cn}

\author{Peng Zhang}
\orcid{0000-0001-9690-7026}
\authornote{Corresponding author.}
\affiliation{
  \institution{Northwestern Polytechnical University}
  \city{Xi'an}
  \state{Shanxi}
  \country{China}
}
\affiliation{
  \institution{Ningbo Institute of Northwestern Polytechnical University}
  \city{Ningbo}
  \state{Zhejiang}
  \country{China}
}
\email{zh0036ng@nwpu.edu.cn}

\author{Wei Huang}
\orcid{0000-0002-0541-8612}
\affiliation{
  \institution{Nanchang University}
  \city{Nanchang}
  \state{Jiangxi}
  \country{China}
}
\email{huangwei@ncu.edu.cn}

\author{Yufei Zha}
\orcid{0000-0001-5013-2501}
\affiliation{
  \institution{Northwestern Polytechnical University}
  \city{Xi'an}
  \state{Shanxi}
  \country{China}
}
\affiliation{
  \institution{Ningbo Institute of Northwestern Polytechnical University}
  \city{Ningbo}
  \state{Zhejiang}
  \country{China}
}
\email{yufeizha@nwpu.edu.cn}

\author{Tao You}
\orcid{0000-0003-0023-7617}
\affiliation{
  \institution{Northwestern Polytechnical University}
  \city{Xi'an}
  \state{Shanxi}
  \country{China}
}
\email{youtao@nwpu.edu.cn}

\author{Yanning Zhang}
\orcid{0000-0002-2977-8057}
\affiliation{
  \institution{Northwestern Polytechnical University}
  \city{Xi'an}
  \state{Shanxi}
  \country{China}
}
\email{ynzhang@nwpu.edu.cn}

\begin{abstract}
Self-supervised sound source localization is usually challenged by the modality inconsistency. In recent studies, contrastive learning based strategies have shown promising to establish such a consistent correspondence between audio and sound sources in visual scenarios. Unfortunately, the insufficient attention to the heterogeneity influence in the different modality features still limits this scheme to be further improved, which also becomes the motivation of our work. In this study, an Induction Network is proposed to bridge the modality gap more effectively. By decoupling the gradients of visual and audio modalities, the discriminative visual representations of sound sources can be learned with the designed Induction Vector in a bootstrap manner, which also enables the audio modality to be aligned with the visual modality consistently. In addition to a visual weighted contrastive loss, an adaptive threshold selection strategy is introduced to enhance the robustness of the Induction Network. Substantial experiments conducted on SoundNet-Flickr and VGG-Sound Source datasets have demonstrated a superior performance compared to other state-of-the-art works in different challenging scenarios. The code is available at \url{https://github.com/Tahy1/AVIN}.
\end{abstract}



\maketitle

\section{Introduction}
Audio-visual Sound Source Localization (AV-SSL) can fundamentally support the intelligent Human-Computer Interaction (HCI) \cite{COWLEY199269} by imitating the perceptive connection of human beings. To bridge the consistency between different modalities for sensing strengthen, the aggregation/combination of distinct modality representations has been employed \cite{ma2021end,ephrat2018looking,arandjelovic2017look} for representation alignment, but the intrinsic disparities in modalities usually limit such a capability of AV-SSL to achieve more robust performance in a variety of scenarios.

Comparatively, the metric learning has been adopted in recent studies to acquire uniform audio-visual representations for modal semantics synchronization, e.g. cosine similarity calculation of audio-visual modalities \cite{hu2020discriminative}, which can benefit the localization of sound sources.
\begin{figure}
  \includegraphics[width=\linewidth]{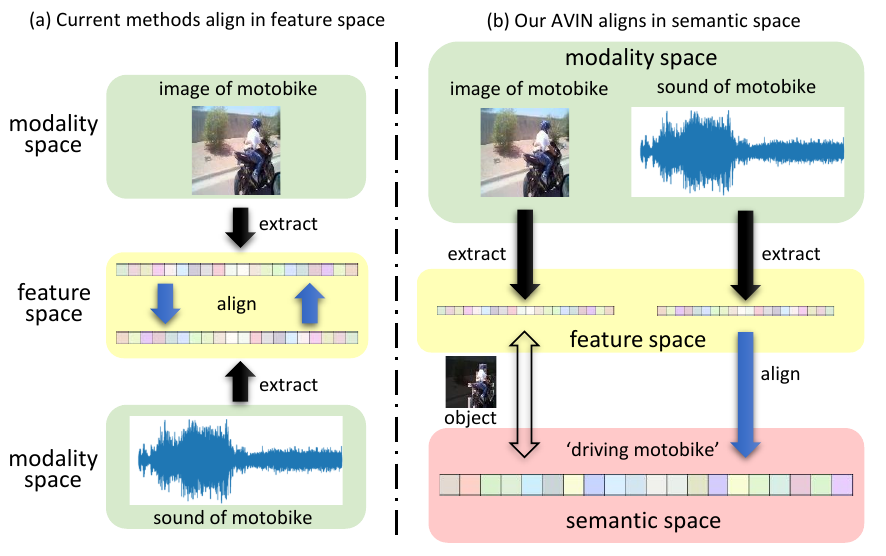}
  \caption{Two approaches of audio-visual representation alignment. (a) Current solutions align audio and visual modalities in feature space directly. (b) Our AVIN extract semantics from visual modality and align with audio modality in semantic space.}
  \label{fig_simple}
\end{figure}
By aligning the audio and visual modalities, some approaches choose to fuse the output features of a modality-specific encoder and minimize an objective function, such as Information Noise Contrastive Estimation (InfoNCE) loss or cross-entropy loss, for a proxy task \cite{zhu2021deep}. A representative work based on modality alinement is proposed by Senocak et al. \cite{senocak2019learning}, which locate sound sources by learning the relationship between audio and global visual features. In a different way, Chen et al \cite{chen2021localizing}. utilizes the pixel-wise audio-visual correspondence to establish the connection between audio and local visual features. With a thresholding operation to categorize the features into positive, negative, and ignore regions, \cite{chen2021localizing} proposes to maximize the cosine similarity of representations between the audio and the region of sound source in visual modality.

However, since the heterogeneity of different modalities has been widely disregarded in current solutions (align in feature space directly, as illustrated in Figure~\ref{fig_simple} (a)), the gap of audio-visual modalities still challenges the connection between the sounding objects with corresponding sound sources. Furthermore, the gradients coupling of distinct modalities also limits an enhancement of performance.

In this work, an Audio-Visual Induction Network (AVIN) is introduced to achieve more effective sound source localization. As illustrated in Figure~\ref{fig_simple} (b), the main operations in the proposed AVIN involves the semantic information extraction from the modality in a low-cost manner, as well as the representations alignment of different modalities based on the semantic information.

Assuming that the sounding objects exist in the image, the proposed Induction Network is capable of adequately exploiting the spatial information of visual modalities. Rather than enforcing the alignment of audio and visual modalities directly, the visual modality is inducted to distill the representation of the complete sounding object in a bootstrapped manner, which is then followed by the alignment of audio modality.

Furthermore, we have discovered that the stop-gradient (stop-grad) can significantly benefit the learning of consistent audio-visual representation. In more recent self-supervised pre-training tasks \cite{grill2020bootstrap,chen2021exploring}, the stop-grad has been widely used to prevent the representation collapse of Siamese networks, but it is unlike the two-stream audio-visual networks that does not typically collapse into a constant. Nonetheless, the coupled gradient of audio-visual modalities during learning still makes the back-propagation process intricate and unstable, and weaken the acquisition of consistent audio-visual representations. To overcome those challenges, the stop-grad is also employed in the proposed work to decouple the gradients of the two modalities, such that the gradient of a particular modal sub-network is autonomous of the other modality. As expectation, the overall model performance can be substantially improved because each sub-network updates parameters independently. Our main contributions can be summarized as follows:

\begin{itemize}
    \item An Audio-Visual Induction Network (AVIN) is proposed to learn a unified audio-visual representation. Based on the extracted visual feature map to generate the semantic representation of the sound source, the obtained Induction Vector can guide the network to project the audio and visual features of objects into a unified semantic space.

    \item The operation of stop-grad is initially introduced into the audio-visual sound source localization task to overcome the gradient coupling of distinct modalities. When the representation of one modality is regarded as constant, the gradient of the other modality can be independently obtained to decouple the gradient between the two modalities.
 
    \item To facilitate the training of the visual network, an adaptive threshold selection strategy is proposed to categorize the similarity score of the visual representation and Induction Vector into foreground, ignore, and background tri-maps, in which the optimal threshold can also be determined accordingly. 

    \item Based on the similarity of visual features between samples as a weight, a visual weighted contrastive loss is designed for training robustness enhancement.
\end{itemize} 
\section{Related Works}
\subsection{Audio-Visual Self-Supervised Representation Learning}
Audio-visual self-supervised representation learning relies on proxy tasks to generate supervised signals \cite{zhu2021deep, ZHANG2022102160}. Modern approaches solve this problem using contrastive learning. Owens et al. \cite{owens2018audio} propose a binary classification approach that considers corresponding audio-visual pairs as positive and asynchronous audio-visual pairs as negative. Korbar et al. \cite{korbar2018cooperative} construct a two-stream network that minimizes the Euclidean distance of audio and visual features with contrastive loss, thereby ensuring that the audio-visual network is semantically coherent and temporally aligned. In contrast, Asano et al. \cite{asano2020labelling} use improved Sinkhorn-Knopp algorithm \cite{cuturi2013sinkhorn} to assign pseudo labels to audio and visual features as supervision signals. Recently, audio-visual representation learning has been treated as an instance discrimination task \cite{morgado2021robust,morgado2021audio,wu2018unsupervised,ye2019unsupervised}, in which the cosine similarity of synchronized audio-visual pairs is maximized through the use of noise contrastive estimation (NCE) or Info NCE loss. Although these works adopt a contrastive learning \cite{oord2018representation, QUAN2022102162} approach to learn audio-visual representations, they do not consider the issue of gradient coupling of distinct modalities. Compared to prior work, the stop-grad is employed in our network to decouple the gradients of the two modalities. By treating the representation of the corresponding modality as constant, the gradient of the current modality is solely related to itself.

\subsection{Audio-Visual Sound Source Localization}
Cognitive science and psychology theories suggest that visual information associated with sound would significantly enhance the searching efficiency in sound space, as demonstrated in \cite{jones1975eye}. In neuroscience, Garner et al. \cite{garner2022cortical} discover that the primary visual cortex can suppress the visual responses after the association between auditory and visual stimuli. These findings have advocated the research interest in the area of audio-visual sound source localization.

Typical sound source localization relies on acoustic hardware. E.g. Zunino et al. \cite{zunino2015seeing} design a device equipped with a microphone array. By integrating the orientation information from the array with the visual information of the camera, the performance of visual tracking can be enhanced, but the main limitation of this scheme is the monophonic sound processing due to the complexity of required hardware. In a different way, other methods rely on spatial sparsity to determine sound source locations. Kidron et al. \cite{kidron2005pixels} utilizes canonical correlation analysis (CCA) to exploit the spatial sparsity of audio-visual events and avoid the issues of dimensionality. Barzelay et al. \cite{barzelay2007harmony} employ instances of significant change within each modality to determine cross-modal associations and visual locations based on handcrafted motion cues.

With the development of deep networks, more effective techniques have been employed in recent works. Some methods \cite{owens2018audio,sharma2020cross,hu2020cross,qian2020multiple} exploit CAM to assist sound source localization. Based on detected object proposal boxes, \cite{truong2021right,parekh2019identify,tian2021cyclic,xuan2022proposal,shi2022unsupervised} determine whether the potential objects are sound sources according to the learned audio feature. By computing the cosine similarity between audio-visual features, recent works \cite{tian2018audio,hu2019deep,senocak2018learning,hu2020discriminative,song2022self,hu2022mix,senocak2022less} take advantage of two-stream network architecture to predict the spatial location of sound sources.

For the solutions above, aligning the data of audio and vision at the feature space is challenging because of the significant differences between the feature spaces of the two modalities, resulting in the difficulty of accurate sound source localization. Instead of aligning audio-visual modalities in feature space directly, this work also takes into account the heterogeneity of audio and vision. The proposed Induction Network performs the alignment in semantic space, which is to ensure consistent semantic and accurate localization results.

\section{Method}
\begin{figure*}
  \includegraphics[width=\textwidth]{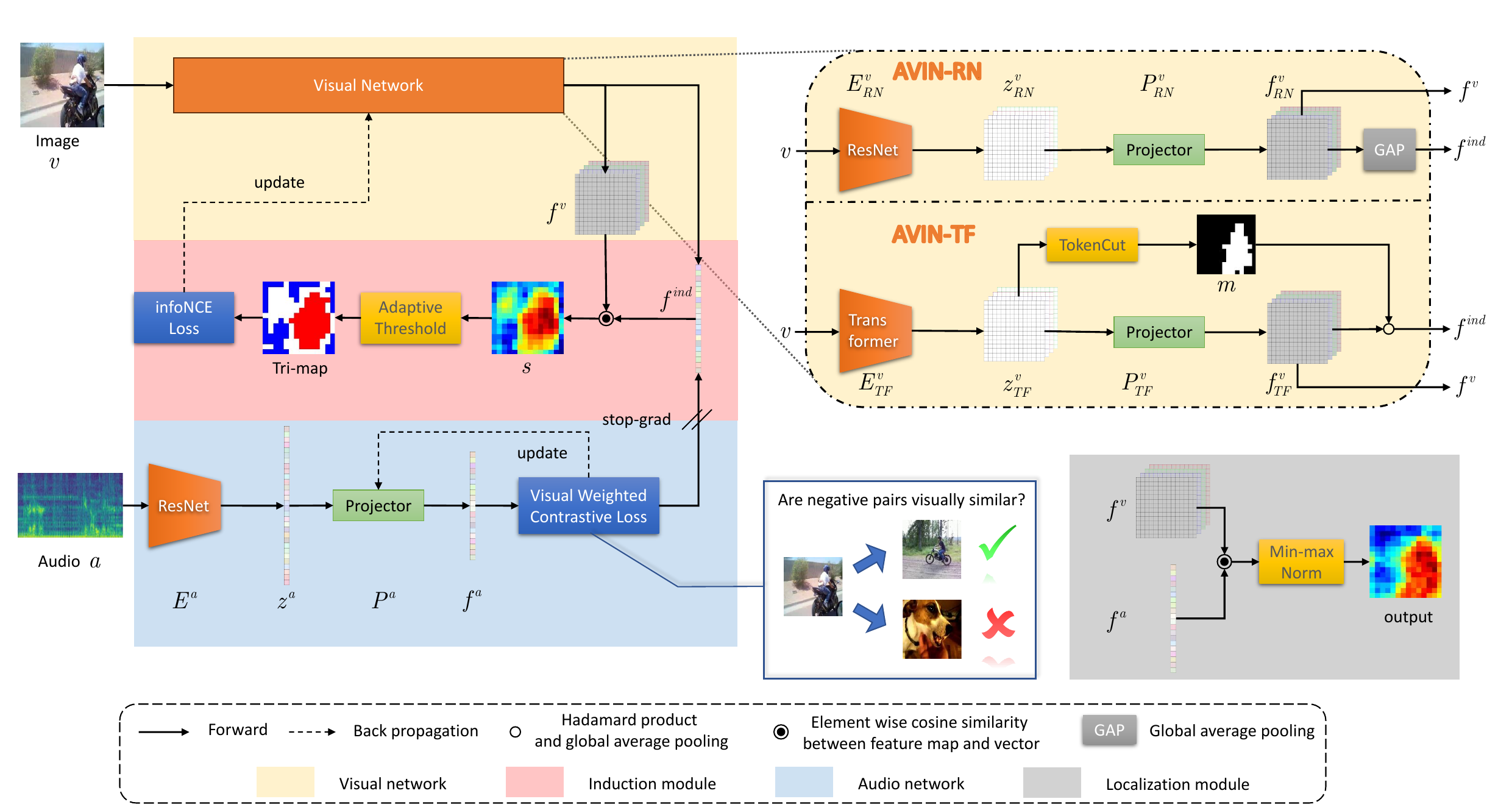}
  \caption{Architecture of Audio-Visual Induction Network. The four parts of the network: visual network, audio network, induction module and localization module are distinguished by different colors. Note that the induction module is only used during training and the localization module is only used during inference.}
  \label{fig_avin}
\end{figure*}
Figure~\ref{fig_avin} presents the overall architecture of the proposed AVIN, and it is a two-stream model with bifurcated audio and visual modalities fused at the bottom. The AVIN is composed of: visual network, audio network, induction module (only for training), and localization module (only for inference). The corresponding operations mainly contain: using the generated visual modality features to obtain Induction Vectors, which act as intermediate vectors to connect the audio and visual modality representations. Then, an adaptive threshold selection strategy is performed with Induction Vector to learn and induct the candidate sounding object regions in the image, which is further to obtain a unified and discriminative representation in the common semantic space. Finally for the audio modality, a visual weighted contrastive loss is designed to align the audio with the visual representation, which is able to avoid faulty negative samples during training phase.

Given a video clip, the central frame $v\in\mathbb{R}^{3\times H\times W}$ together with a 3s audio log-mel spectrogram $a\in\mathbb{R}^{1\times F\times T}$ are input into the network, in which $H$ and $W$ denote the height and width of the frame, $F$ represents the number of mel-frequency bins of the spectrogram, and $T$ is the number of audio frames. The functionality of each part in AVIN is elaborated as below.

\subsection{Visual Network}
The visual network consists of a Projector $P^v$ and a Visual Encoder $E^v$, which is formed by the ResNet or Transformer alternatively.

\noindent\textbf{ResNet}: As in \cite{chen2021localizing}, ResNet18 is employed as our visual encoder, which consists of 8 residual blocks \cite{he2016deep}. In order to maintain the spatial information of the output, the average pooling layer and the fully connected layer are excluded at the end of the network. The visual embedding is denoted as $z_{RN}^v\in\mathbb{R}^{c_{RN}^v\times h\times w}$, where $c_v$ represents the number of channels of the feature. The $h=\left\lfloor\frac{H}{16}\right\rfloor$ and $w=\left\lfloor\frac{W}{16}\right\rfloor$ denote the height and width of $z_{RN}^v$, respectively.

\noindent\textbf{Transformer}: The Vision Transformer \cite{dosovitskiy2020image} reshapes the image into a sequence of non-overlapping patch sets $v^\prime\in\mathbb{R}^{n\times(p^2\times3)}$, where $p \times p$ is the resolution of each patch, and $n=HW/p^2$ represents the number of patches. By concatenating a learnable CLS token before the first patch, a total of $(n+1)$ tokens can be obtained. A positional embedding is then added to the token before feeding into the Transformer Encoder \cite{vaswani2017attention}, which consists of 12 Transformer blocks. Each block is composed of a Multi-Layer Perception (MLP), Muti-Head Self-Attention (MHSA) layer, and a normalization layer. In MHSA, each token is projected into a query Q, key K, and value V, the attention between tokens is computed using:
\begin{equation}
  \text{Attention}\left(Q,K,V\right)=\text{softmax}\left(\frac{QK^T}{\sqrt{d_k}}\right)V \label{eq_qkv}
\end{equation}
where $d_k$ is the dimension of the hidden layer in MHSA. In accordance with TokenCut \cite{wang2022tokencut}, the weights of the self-supervised trained DINO \cite{caron2021emerging} are utilized to initialize the Vision Transformer. The key of the last MHSA layer in the final block is used as the visual embedding $z_{TF}^v\in\mathbb{R}^{c_{TF}^v\times h\times w}$ of the transformer, and the CLS token is discarded.

\noindent\textbf{Visual Projector}: To align audio and visual embeddings, a projector $P^v$ is proposed to project the embeddings into a common space $f^v=P^v\left(z^v\right)$, $f^v\in\mathbb{R}^{c\times h\times w}$ with dimension $c$. In our work, distinct projectors have been employed for different network architectures, with a convolutional layer (Conv) utilized for ResNet, as well as a Conv-ReLU-Conv configuration for transformer. The reason behind using different projectors for different backbones is that the parameters of ResNet are updated during  training in comparison to the fixed parameters of transformer. Hence, additional nonlinear and convolutional layers can be incorporated to augment the expressive capabilities.

\subsection{Audio Network}
Similar to the vision network, the audio network consists of an audio encoder and an audio projector as well.

\noindent\textbf{Audio Encoder}: In this work, ResNet22 pre-trained by PANNs \cite{kong2020panns} acts as the audio encoder, which consists of 8 residual blocks, 4 supplementary convolutional layers, and two fully connected (FC) layers. To generate audio embeddings, the last classification layer is removed. The output of the audio encoder is symbolized as $z^a\in\mathbb{R}^{c^a\times1}$, where $c^a$ denotes the number of channels in the audio embedding.

\noindent\textbf{Audio Projector}: The projector is employed to project audio embeddings into a common space for the computation of similarity between audio and visual representations. Since audio embeddings are represented as one-dimensional vectors, we employ the FC-ReLU-FC structure to derive the audio representation $f^a=P^a\left(z^a\right)$, $f^a\in\mathbb{R}^{c\times1}$, with an equal number of channels to the visual representation.

\subsection{Induction Module}
The audio and visual representations are connected in induction module with an intermediate vector `Induction Vector', defined as $f^{ind}$, which is obtained from visual modality and is supposed to represent the semantics of the sound source. For visual networks, $f^{ind}$ is utilized to induct $f^v$ to acquire more precise object representations through a bootstrapping fashion. For audio networks, $f^a$ can be aligned with $f^{ind}$ using a visual weighted contrastive loss.
~\\
\subsubsection{\textbf{Induction Vector Generation}}
\begin{figure}
  \includegraphics[scale=0.4]{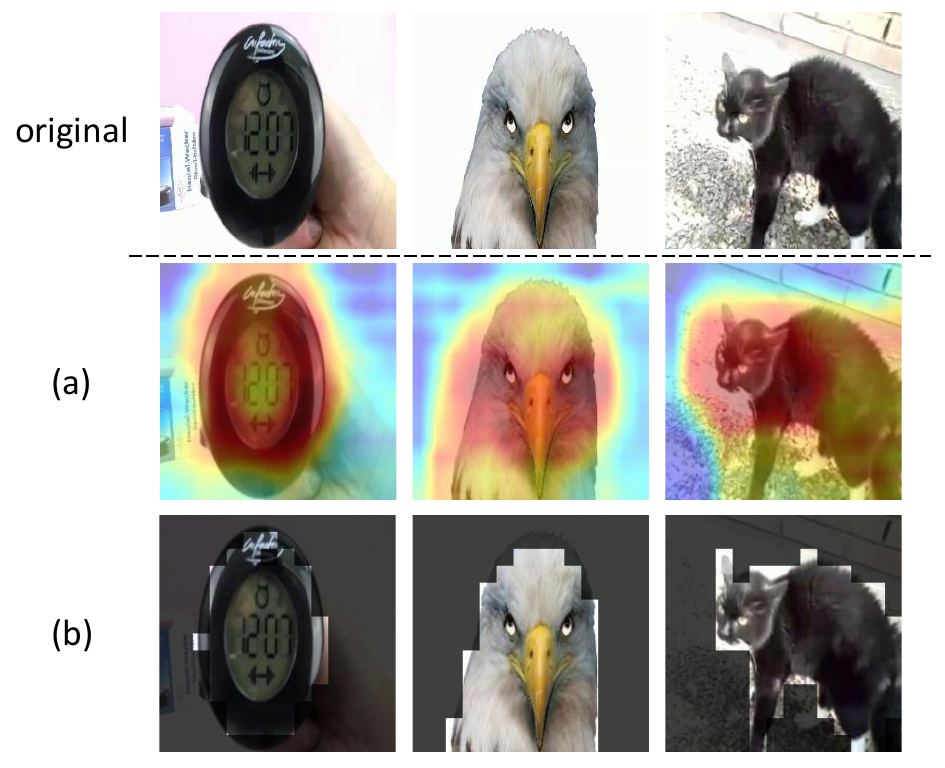}
  \caption{Visualization of salient region. (a) Pixel-wise cosine similarity score map of ResNet features after global pooling on the feature map. (b) Foreground mask extracted by TokenCut for transformer output features.}
  \label{fig_region}
\end{figure}

~\\
To obtain $f^{ind}$ from $f^v$, global average pooling (GAP) is performed in visual network based on ResNet as backbone:
\begin{equation}
  f^{ind}=\text{GAP}(f^v)
\end{equation}
Considering that the ResNet classification network projects the pooled features linearly into logits in the category space, the pooled feature is further utilized based on the guiding intuition that contains a specific category of semantic information. For validation, the cosine similarity is calculated between $f^{ind}$ and the visual representation map $f^v\left(i,j\right)$ at spatial location $\left(i,j\right)$ using Equation~\ref{eq_svv}:
\begin{equation}
  s^{vv}\left(i,j\right)=\frac{\left<f^v(i,j),f^{ind}\right>}{\left\|f^v(i,j)\right\|_2\left\|f^{ind}\right\|_2},(i,j)\in[h]\times[w] \label{eq_svv}
\end{equation}
where $\left<\cdot,\cdot\right>$ denotes the inner product, $s^{vv}\in\mathbb{R}^{h\times w}$. Figure~\ref{fig_region} (a) depicts the visualization outputs of $s^{vv}$. The foreground region in the image has a high score, which indicates that the pooled feature $f^{ind}$ has a strong similarity to the foreground. Based on the assumption that the existence of sounding objects in image, the target representation from pre-training contained in $f^{ind}$ can serve as the semantics of the sound source.

For visual networks using Transformer as the backbone, an unsupervised object detection method TokenCut \cite{wang2022tokencut} is employed to generate $f^{ind}$. Based on the finding of DINO (self-distillation with no labels trained vision transformer features contain information of the semantic segmentation), TokenCut adopts Normalized Cut (NCut) \cite{shi2000normalized} to divide Key features from the last attention layer of the DINO-pretrained model into two sets, foreground and background, as shown in Figure~\ref{fig_region} (b). To obtain the foreground, the similarity matrix $M$ of $z_v^{TF}$ is computed based on the Equation~\ref{eq_Ms} as:
\begin{equation}
  M_s(i,j)=\frac{\left<z_i,z_j\right>}{\left\|z_i\right\|_2\left\|z_j\right\|_2},z_i,z_j\in\{z^{TF}_v\} \label{eq_Ms}
\end{equation}
where $z_i$ and $z_j$ represent tokens in $z_{TF}^v$, $M$ is an $n\times n$ symmetrical matrix. $M_b$ is derived after binarizing $M$ as:
\begin{equation}
  M_b(i,j)=\begin{cases}
    1 &M(i,j)\geq\tau_m \\
    \epsilon &M(i,j)<\tau_m
  \end{cases}
\end{equation}
where $\tau_m=0.2$ and $\epsilon={10}^{-5}$ are the hyperparameters. The set composed of $z_{TF}^v$ is denoted as $\mathcal{Z}=\{\text{z}_i\}$, with each element $\text{z}_i$ as the node, and the similarity between two nodes as the edge denoted by $\mathcal{M}$. A fully connected undirected graph $\mathcal{G}=(\mathcal{Z},\mathcal{M})$ can be constructed with the edges and nodes, and each node is linked to others by edges. To divide the graph into two disjoint sets $\mathcal{A}$ and $\mathcal{B}$, NCut performs the minimization as:
\begin{equation}
  \frac{C(\mathcal{A},\mathcal{B})}{C(\mathcal{A},\mathcal{Z})}+\frac{C(\mathcal{A},\mathcal{B})}{C(\mathcal{B},\mathcal{Z})}
\end{equation}
where $C(\cdot,\cdot)$ represents the sum of edges between the nodes within the two sets. According to \cite{shi2000normalized}, solving a generalized eigensystem $\left(D-M_b\right)y=\lambda Dy$ enables the discovery of the second smallest eigenvector $y_1$, where $D$ is:
\begin{equation}
  D(i,j)=\begin{cases}
    \sum_jM_b(i,j) &i=j \\
    0 &i\neq j
  \end{cases}
\end{equation}
Then the divided sets are $\mathcal{A}=\{z_i|y_1^i\le\bar{y_1}\}$ and $\mathcal{B}=\{z_i|y_1^i>\bar{y_1}\}$, where $\bar{y_1}$ is the average of $y_1$. The set of tokens with the largest absolute value of the eigenvalue is denoted as the foreground set $\mathcal{F}$, and the foreground mask $m\in\mathbb{R}^n$ is obtained by
\begin{equation}
  m_i=\begin{cases}
    1 &z_i\in\mathcal{F} \\
    0 &z_i\notin\mathcal{F}
  \end{cases}
  \label{eq_mask}
\end{equation}
where $m$ is reshaped to $1\times h\times w$, and the Induction Vector $f^{ind}$ can be generated by
\begin{equation}
  f^{ind}=\text{GAP}(f_v\circ m)
\end{equation}
where $\circ$ is the Hadamard product.
~\\
\subsubsection{\textbf{Bootstrapped Induct Visual Network}}
~\\
The training of the visual network has two objectives: (1) to project the region in the visual representation $f_v$, which semantically corresponds to the sound source, into a unified representation distinct from other semantics, and (2) to generate a high-quality Induction Vector $f^{ind}$ from the semantically explicit visual representation $f^v$.

\begin{figure}
  \includegraphics[width=\linewidth]{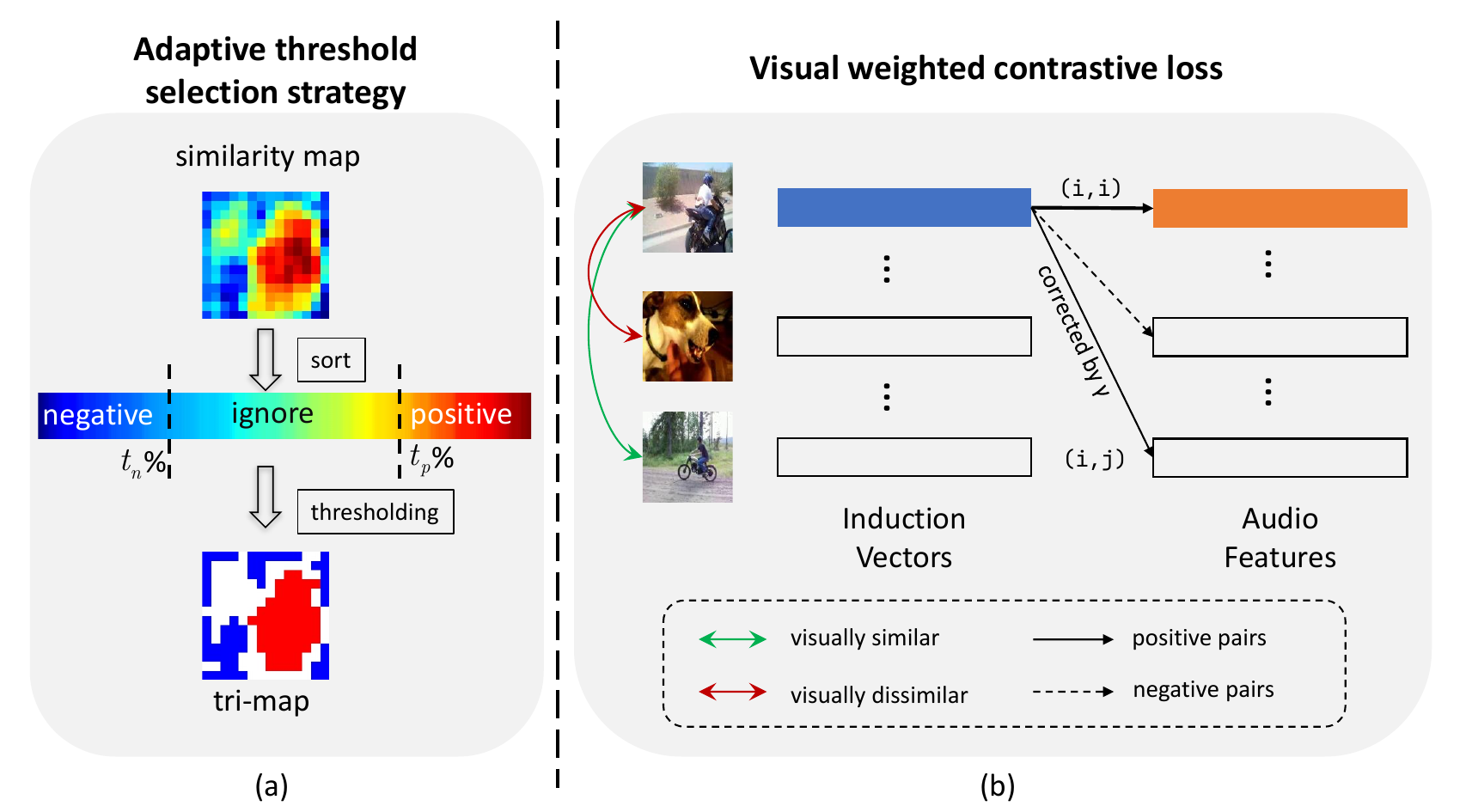}
  \caption{Schematic of (a) adaptive threshold selection strategy and (b) visual weighted contrastive loss.}
  \label{fig_two_modules}
\end{figure}

In this work, we employ the tri-map strategy in HardWay \cite{chen2021localizing} to train the visual network. The proposed AVIN differs from HardWay in two aspects: 1) AVIN uses the Induction Vector to generate a similarity map for visual representation, while HardWay utilizes the audio representation; 2) A proposed adaptive threshold selection strategy is employed to  obtain the tri-map as shown in Figure~\ref{fig_two_modules} (a), while HardWay employs a fixed threshold. 

The similarity map $s^{vv}$ in Equation~\ref{eq_svv} is used to separate the image into positive, ignore, and negative regions, i.e., tri-map. It is pre-defined in AVIN that a certain percentage is attributed to the foreground region $t_p\%$ for each image, while $t_n\%$ is assigned to the background region, and the remaining areas are ignored. Specifically, the scores of $h\times w$ in the similarity map $s^{vv}$ are sorted in ascending order. The minimum value of the top $t_p\%$ of the scores is denoted as $\epsilon_p$, while the maximum value of the bottom $t_n\%$ scores is denoted as $\epsilon_n$.
\begin{equation}
  \begin{aligned}
    & {\hat{m}}_{ijp}=\text{sigmoid}\left(\left(S_{ij}-\epsilon_p\right)/\tau_s\right) \\
    & {\hat{m}}_{ijn}=1-\text{sigmoid}\left(\left(S_{ij}-\epsilon_n\right)/\tau_s\right) \\
    & {SP}_{ij}=\frac{\left<{\hat{m}}_{ijp},S_{ij}\right>_F}{\left|{\hat{m}}_{ijp}\right|} \\
    & {SN}_{ij}=\frac{\left<{\hat{m}}_{ijn},S_{ij}\right>_F}{\left|{\hat{m}}_{ijn}\right|}
  \end{aligned}
  \label{eq_hardway}
\end{equation}
where $\hat{m}$ is the soft mask, $\left<\cdot,\cdot\right>_F$ is the Frobenius inner product, $\tau_s=0.03$ is the hyperparameter that controls the degree of softening, and $N$ represents the number of samples in a batch. $S_{ij}$ denotes the cosine similarity map between the $j$-th Induction Vector $f^{ind}$ and the visual representation map $f^v$ of the $i$-th sample. For the training of visual network, infoNCE is adopted as the loss function:
\begin{equation}
  \mathcal{L}_v=-\frac{1}{N}\sum_{i=1}^{N}\left[\log{\frac{\exp\left({SP}_{ii}/\tau_c\right)}{\sum_{j}\exp{\left({SP}_{ij}/\tau_c\right)}+\sum_{j}\exp{\left({SN}_{ij}/\tau_c\right)}}}\right]
\end{equation}
with $\tau_c=0.07$ as the temperature hyperparameter of infoNCE.
~\\
\subsubsection{\textbf{Audio Representation Learning}}
~\\
To establish a correlation between audio and visual representations, the Induction Vector serves as a bridge to connect the two modalities. 
As shown in Figure~\ref{fig_two_modules} (b), a visual weighted contrastive loss is introduced to facilitate learning the projection of audio representation in a common space.
\begin{equation}
  \begin{aligned}
    & \text{Eu}_{ij} = d\left(f_i^{ind},f_j^a\right)^2 \\
    & \gamma_{ij}=-\text{cossim}\left(f_i^{ind},f_j^{ind}\right) \\
    & \mathcal{L}_a=\frac{1}{N}\sum_{i=1}^{N}{\max\left(0,\text{Eu}_{ii}-\frac{1}{N-1}\sum_{j=1}^{N}{1\left(i\neq j\right)\cdot\gamma_{ij}\cdot\text{Eu}_{ij}}+\theta\right)}
  \end{aligned}
  \label{eq_la}
\end{equation}
where $d\left(\cdot,\cdot\right)$ denotes the Euclidean distance between two vectors, the margin hyperparameter $\theta=0.6$ denotes the minimum margin between positive and negative samples. $\gamma_{ij}$ represents the negative cosine similarity of the Induction Vector between two distinct samples, and it controls the direction, as well as the scale of the negative samples in the contrastive loss. It is noteworthy that $f^{ind}$ is generated by the visual modality for the representation of object in the image. If the visual similarity between two samples is high, i.e., $\gamma<0$, their sound is supposed to be similar as well, and it indicates that the $j$-th sample should be classified as a positive sample. On the other hand, if the visual similarity between two samples is low, i.e., $\gamma\geq0$, it means that they are likely to be different objects and sound different either. Thus, the $j$-th and $i$-th samples are classified as negative sample pairs with the weight $\gamma$ accordingly.
~\\
\subsubsection{\textbf{Stop-Grad Consideration}}
~\\
The loss function of AVIN is defined as:

\begin{equation}
  \mathcal{L}=\mathcal{L}_v+\mathcal{L}_a \label{eq_loss}
\end{equation}

During the training phase, the gradients of the visual and audio networks are decoupled. Specifically, for $\mathcal{L}_a$ as given by Equation~\ref{eq_la}, all vectors associated with the visual modality, including $f^{ind}$ and $\gamma$, are regarded as constants. A reasonable explanation is that the intricate mutual coupling of gradients may destabilize the backpropagation process, which is caused by the representation differences between the audio and visual modalities. Decoupling the gradients of the two modalities ensures that the parameter update process of the visual network would only relate to the visual modality itself. The audio network regards the Induction Vector from the visual modality as constant, and seeks to minimize the Euclidean distance between the synchronized audio representation and the Induction Vector.

\subsection{Localization Module}
During the inference stage, the localization module is to ascertain the similarity between the audio representation and the visual representation map. Given the audio representation $f^a$ and the visual representation map $f^v$ in the learned common space, the similarity map $s^{av}$ is:
\begin{equation}
  s^{av}\left(i,j\right)=\frac{\left<f^v(i,j),f^{a}\right>}{\left\|f^v(i,j)\right\|_2\left\|f^{a}\right\|_2},(i,j)\in[h]\times[w]
\end{equation}

Finally, a min-max normalization process is performed to re-scale $s^{av}$ to the interval $[0,1]$:
\begin{equation}
  {\widetilde{s}}^{av}=\frac{s^{av}-\min\left(s^{av}\right)}{\max{\left(s^{av}\right)}-\min\left(s^{av}\right)}
\end{equation}
where ${\widetilde{s}}^{av}$ represents the output of AVIN and denotes the degree of correlation between the location of each image and the provided audio cues.

\section{Experiments}
\subsection{Comparisons with State-of-the-art Methods}
\begin{table}
    \caption{Sound source localization result on Flickr test set and VGG-SS benchmark}
    \resizebox{\columnwidth}{!}{
        \begin{tabular}{lccccc}
        \toprule
        \multicolumn{1}{c}{\multirow{2}[4]{*}{Method}} & \multirow{2}[4]{*}{Training set} & \multicolumn{2}{c}{Flickr test set} & \multicolumn{2}{c}{VGG-SS} \\
        \cmidrule{3-6}          &       & cIoU  & AUC   & cIoU  & AUC \\
        \midrule
        HardWay \cite{chen2021localizing} & \multirow{6}[2]{*}{Flickr 10k} & 0.582  & 0.525  & 0.288  & 0.351  \\
        SSPL \cite{song2022self} &       & 0.743  & 0.587  & 0.208  & 0.300  \\
        FNAC \cite{sun2023learning} &       & 0.843  & 0.633  & 0.336  & 0.372  \\
        FNAC+OGL \cite{sun2023learning} &       & 0.847  & 0.643  & 0.407  & 0.404  \\
        AVIN-RN &       & \textbf{0.868} & \textbf{0.659} & \textbf{0.423}  & \textbf{0.421}  \\
        AVIN-TF &       & 0.843  & 0.632  & 0.413 & 0.418 \\
        \midrule
        HardWay \cite{chen2021localizing} & \multirow{6}[2]{*}{Flickr 144k} & 0.699  & 0.573  & 0.269  & 0.344  \\
        SSPL \cite{song2022self} &       & 0.759  & 0.610  & 0.289  & 0.356  \\
        FNAC \cite{sun2023learning} &       & 0.787  & 0.593  & 0.348  & 0.380  \\
        FNAC+OGL \cite{sun2023learning} &       & 0.840  & 0.631  & 0.406  & 0.403  \\
        AVIN-RN &       & \textbf{0.872} & \textbf{0.658} & \textbf{0.423}  & \textbf{0.420}  \\
        AVIN-TF &       & 0.843  & 0.639  & 0.422 & 0.419 \\
        \midrule
        HardWay \cite{chen2021localizing} & \multirow{6}[2]{*}{VGGSound 10k} & 0.618  & 0.536  & 0.291  & 0.368  \\
        SSPL \cite{song2022self} &       & 0.763  & 0.591  & 0.316  & 0.374  \\
        FNAC \cite{sun2023learning} &       & 0.857  & 0.637  & 0.372  & 0.388  \\
        FNAC+OGL \cite{sun2023learning} &       & 0.821  & 0.636  & 0.415  & 0.408  \\
        AVIN-RN &       & \textbf{0.884} & \textbf{0.659} & \textbf{0.450}  & \textbf{0.431}  \\
        AVIN-TF &       & 0.835  & 0.643  & 0.448 & 0.433 \\
        \midrule
        HardWay \cite{chen2021localizing} & \multirow{6}[2]{*}{VGGSound 144k} & 0.719  & 0.582  & 0.292  & 0.367  \\
        SSPL \cite{song2022self} &       & 0.767  & 0.605  & 0.323  & 0.376  \\
        FNAC \cite{sun2023learning} &       & 0.847  & 0.638  & 0.406  & 0.405  \\
        FNAC+OGL \cite{sun2023learning} &       & 0.851  & 0.643  & 0.421  & 0.412  \\
        AVIN-RN &       & \textbf{0.876} & \textbf{0.658} & \textbf{0.449}  & \textbf{0.436}  \\
        AVIN-TF &       & 0.851  & 0.644  & 0.448 & 0.433 \\
        \bottomrule
    \end{tabular}
    \label{tab_sota}
    }
\end{table}
The proposed AVIN is firstly compared with other works on the SoundNet-Flickr test set as shown in Table~\ref{tab_sota}. We employ ResNet and Transformer as visual encoders denoted as AVIN-RN and AVIN-TF, respectively. When training on Flickr 10k and 144k, AVIN-TF is comparable to the recently proposed FNAC \cite{sun2023learning}, while AVIN-RN outperforms the previous best (0.868 vs. 0.847 in 10k and 0.872 vs. 0.840 in 144k). Noticed that FNAC+OGL incorporates Object-Guided Localization (OGL), which is a post-processing strategy to refine localization results. In comparison, the output of AVIN is only based on the correspondence of audio and visual features, but achieves superior results. For cross-dataset evaluation purposes, the proposed models are trained on the VGGSound 10k and 144k training sets. Since a greater diversity of video categories is presented in VGGSound compared to SoundNet-Flickr, the AVIN can effectively establish the association between visual and audio using the induction vector, which has achieved state-of-the-art results and validated the cross-dataset generalization ability in both settings.

The evaluation results on VGG-SS benchmark are also illustrated in Table~\ref{tab_sota}. With the multiple reproductions, the best results are reported because the sample count in the test set is less than \cite{chen2021localizing} (4664 vs. 5158). AVIN surpasses all the other works with a clear margin. In the VGGSound 10k training case, AVIN-RN outperforms FNAC+OGL by 8.4\% cIoU and 5.6\% AUC performance increase, with cIoU of 0.450 and AUC of 0.431. All the results demonstrate the superior performance of the proposed work compared to state-of-the-art works on both datasets.

Furthermore, we observe that smaller subsets (Flickr or VGGSound 10k) exhibit comparable performance to larger ones (Flickr or VGGSound 144k). We hypothesize that AVIN can learn sufficient audio-visual semantic information for satisfactory fitting from a smaller subset. In contrast, the larger subset not only provides less additional semantic information based on the smaller subset but also may affected by overfitting \cite{NEURIPS2022_f3f2ff95}, as evidenced in the AVIN-RN model's performance on the VGGSound dataset in Table~\ref{tab_sota}.

The visualized results for sound source localization of AVIN-RN and AVIN-TF on Flickr test set and VGG-SS are shown in Appendix~\ref{vis} Figure~\ref{fig_visualization}. Our AVIN demonstrates enhanced prediction in localizing the semantic region of the sound source compared to prior works while minimizing background interference. Notably, AVIN-TF exhibits a certain capability to delineate the contours of sound sources.

\subsection{Ablation Study}
In the ablation study conducted in the subsequent experiments, Flickr refers to using SoundNet-Flickr 10k as the training set and evaluating on the SoundNet-Flickr test set, while VGG-SS refers to using VGGSound 10k as the training set and evaluating on VGG-SS.
\begin{table}
    \caption{Ablation study for the induction vector}
    \begin{tabular}{ccccc}
      \toprule
       & Method & Dataset & cIoU  & AUC \\
      \midrule
      \multirow{4}[2]{*}{(a)} & \multirow{2}[1]{*}{AVIN-RN} & Flickr & 0.659  & 0.560  \\
            &       & VGGSound & 0.361  & 0.395  \\
            & \multirow{2}[1]{*}{AVIN-TF} & Flickr & 0.337  & 0.436  \\
            &       & VGGSound & 0.098  & 0.251  \\
      \midrule
      \multirow{4}[2]{*}{(b)} & \multirow{2}[1]{*}{AVIN-RN} & Flickr & 0.522  & 0.491  \\
            &       & VGGSound & 0.263  & 0.337  \\
            & \multirow{2}[1]{*}{AVIN-TF} & Flickr & 0.542  & 0.505  \\
            &       & VGGSound & 0.255  & 0.338  \\
      \bottomrule
    \end{tabular}
    \label{tab_abl_ind}
\end{table}
\subsubsection{\textbf{Induction Vector}}
~\\
To evaluate the contribution of the induction vector, two sets of experiments are conducted: 
(a) remove the induction vector and visual weighted contrastive loss, directly compute the cosine similarity between the visual representation map $f^v$ and the audio vector $f^a$, which means to use the output of the localization module $s^{av}$ instead of $s^{vv}$ to generate the tri-map; (b) retain $f^{ind}$ but remove $\mathcal{L}_v$. Table~\ref{tab_abl_ind} shows the results of the experiments above. In case (a), when $f^{ind}$ is removed, the cIoU performance of AVIN-TF drops to 0.337 (on Flickr) and 0.098(on VGG-SS), while AVIN-RN drops to 0.659(on Flickr) and 0.361(on VGG-SS). A reasonable explanation is that the location of salient objects contained in the pre-trained ResNet features can facilitate the network to determine rough audio-visual correspondence, and in accompany with the induction vector to benefit AVIN-RN to learn precise correspondence. Due to the lack of salient object information, the result of the AVIN-TF is poor after removing the induction vector. In case (b), the performance of AVIN-RN and AVIN-TF drops to similar levels because of insufficient training of bootstrapped visual network, which further validates an informative deficiency without training visual network.
~\\
\subsubsection{\textbf{Stop-Grad Operation}}
~\\
Experiments are conducted to investigate the influence of stop-grad operation on training as shown in Table~\ref{tab_abl_sg_w}. It can be found that AVIN-RN is more sensitive to the gradient than AVIN-TF, and the cIoU performance drops by 0.559 on Flickr and 0.273 on VGG-SS, while AVIN-TF has a drop of 0.092 on Flickr and 0.081 on VGG-SS. The reason is that the gradient updates of the visual and audio networks mainly depend on both modalities' representations simultaneously, which may weaken the training process due to the isomerism of modality. By allowing the visual/audio network to update its parameters only according to the gradient of the visual/audio modality, the stop-grad operation can help to improve the performance of audio-visual sound source localization.

To verify the effect of stop-grad on other audio-visual sound source localization architectures, a HardWay \cite{chen2021localizing} variant is used with the adaptive threshold selection strategy. The experiments are conducted by replacing the audio network with a pretrained ResNet22 model with fixed parameters \cite{kong2020panns}. The visual projector uses a convolutional layer and the audio projector uses a fully connected (FC) layer with the same number of channels. An FNAC variant is also conducted by replacing the audio network with fixed ResNet22. The experimental results are shown in Table~\ref{tab_sg_hardway}, which have verified that the performance is still acceptable with stop-grad even the audio network parameters are fixed. When the audio representation has a gradient and the audio network parameters are updated with training, the performance decreases significantly in comparison to stop-grad. The results indicates that the stop-grad is an effective plug-and-play operation to improve the performance of other architectures, and also promising to benefit the design of future audio-visual networks.
\begin{table}
    \caption{Ablation study for stop-grad and visual weighted contrastive loss}
    \resizebox{\columnwidth}{!}{
    \begin{tabular}{ccccccc}
        \toprule
        \multirow{2}[3]{*}{Stop-grad} & \multirow{2}[3]{*}{Weighted} & \multirow{2}[3]{*}{Method} & \multicolumn{2}{c}{Flickr} & \multicolumn{2}{c}{VGGSound} \\
        \cmidrule{4-7}          &       &       & cIoU  & AUC   & cIoU  & AUC \\
        \midrule
        \multirow{2}[0]{*}{$\times$} & \multirow{2}[0]{*}{\checkmark} & \cellcolor[rgb]{ .91,  .91,  .91}AVIN-RN & \cellcolor[rgb]{ .91,  .91,  .91}0.309  & \cellcolor[rgb]{ .91,  .91,  .91}0.411  & \cellcolor[rgb]{ .91,  .91,  .91}0.177  & \cellcolor[rgb]{ .91,  .91,  .91}0.300  \\
        &       & \cellcolor[rgb]{ .91,  .91,  .91}AVIN-TF & \cellcolor[rgb]{ .91,  .91,  .91}0.751  & \cellcolor[rgb]{ .91,  .91,  .91}0.578  & \cellcolor[rgb]{ .91,  .91,  .91}0.367  & \cellcolor[rgb]{ .91,  .91,  .91}0.397  \\
        \multirow{2}[0]{*}{\checkmark} & \multirow{2}[0]{*}{$\times$} & AVIN-RN & 0.851  & 0.650  & 0.436  & 0.431  \\
        &       & AVIN-TF & 0.755  & 0.602  & 0.437  & 0.419  \\
        \multirow{2}[1]{*}{\checkmark} & \multirow{2}[1]{*}{\checkmark} & \cellcolor[rgb]{ .91,  .91,  .91}AVIN-RN & \cellcolor[rgb]{ .91,  .91,  .91}0.868  & \cellcolor[rgb]{ .91,  .91,  .91}0.659  & \cellcolor[rgb]{ .91,  .91,  .91}0.450  & \cellcolor[rgb]{ .91,  .91,  .91}0.431  \\
        &       & \cellcolor[rgb]{ .91,  .91,  .91}AVIN-TF & \cellcolor[rgb]{ .91,  .91,  .91}0.843  & \cellcolor[rgb]{ .91,  .91,  .91}0.632  & \cellcolor[rgb]{ .91,  .91,  .91}0.448  & \cellcolor[rgb]{ .91,  .91,  .91}0.433  \\
        \bottomrule
    \end{tabular}
    \label{tab_abl_sg_w}
    }
\end{table}
\begin{table}
    \caption{Stop-grad on HardWay \cite{chen2021localizing} and FNAC \cite{sun2023learning} variant}
    \begin{tabular}{cccccc}
    \toprule
          \multirow{2}[4]{*}{Method} & \multirow{2}[4]{*}{stop-grad} & \multicolumn{2}{c}{Flickr} & \multicolumn{2}{c}{VGGSound} \\
\cmidrule{3-6}          &       & cIoU  & AUC   & cIoU  & AUC \\
    \midrule
    \multirow{2}[2]{*}{Hardway \cite{chen2021localizing}} & $\times$     & 0.731  & 0.583  & 0.401  & 0.414  \\
          & \checkmark     & 0.811  & 0.614  & 0.412  & 0.419  \\
    \midrule
    \multirow{2}[2]{*}{FNAC \cite{sun2023learning}} & $\times$     & 0.755  & 0.588  & 0.368  & 0.389  \\
          & \checkmark     & 0.779  & 0.608  & 0.372  & 0.390  \\
    \bottomrule
    \end{tabular}
    \label{tab_sg_hardway}
\end{table}
~\\
\subsubsection{\textbf{Visual Weighted Contrastive Loss}}
~\\
To validate the robustness of visual weights for AVIN, the visual weight parameter $\gamma$ is set to 1 with the vanilla contrastive loss to train the audio network. The results presented in Table~\ref{tab_abl_sg_w} (b) indicate that the performance of AVIN for both two architectures drops to different degrees. Since the vanilla contrastive loss cannot distinguish between audio-visual pairs with the same semantics contained in negative pairs, the Euclidean distance between $f^{ind}$ and $f^a$ is incorrectly maximized. Comparatively, the proposed visual weighted contrastive loss, which is based on visual priors, can correct and weight the erroneous negative pairs as positive pairs, thereby substantially enhance the overall robustness.
~\\
\subsubsection{\textbf{Percentage of Adaptive Threshold}}
~\\
The performance of network is affected by the hyperparameter of adaptive threshold selection strategy. As a solution, we train AVIN-RN and AVIN-TF with different $t_p$ and $t_n$ combinations respectively, and the results are shown in Table~\ref{tab_abl_tptn}. For AVIN-RN, the results show less susceptibility to the ratios, and a broad range of thresholding ratios (20-40\% for $t_p$, 10-50\% for $t_n$) yield satisfactory outcomes. However, AVIN-TF is more sensitive to $t_p$ and can achieve better results when $t_p$ is 20-30\%.

Additionally, we replace the tri-map with the bi-map generated by TokenCut, which involves reshaping the mask $m$ in Equation~\ref{eq_mask} to the scale of $h\times w$, where ${\hat{m}}_{iip}=m$, ${\hat{m}}_{iin}=\mathbf{1}-m$, ${\hat{m}}_{ijn}=\mathbf{1}$ in Equation~\ref{eq_hardway}, and $\mathbf{1}$ denotes a $h\times w$ tensor of all ones. Compared to the performance of bi-map in last row of Table~\ref{tab_abl_tptn}, the tri-map generated by the adaptive threshold selection strategy can achieve better performance.
\begin{table}
    \caption{Ablation study for adaptive threshold}
    \begin{tabular}{ccccccc}
        \toprule
        \multirow{2}[4]{*}{Method} & \multirow{2}[4]{*}{$t_p$ (\%)} & \multirow{2}[4]{*}{$t_n$ (\%)} & \multicolumn{2}{c}{Flickr} & \multicolumn{2}{c}{VGGSound} \\
        \cmidrule{4-7}          &       &       & cIoU  & AUC   & cIoU  & AUC \\
        \midrule
        \multirow{9}[2]{*}{AVIN-RN} & 10    & 30    & 0.843  & 0.637  & 0.414  & 0.420  \\
        & 10    & 50    & 0.847  & 0.636  & 0.417  & 0.420  \\
        & 20    & 10    & 0.868  & 0.650  & 0.429  & 0.426  \\
        & 20    & 30    & \textbf{0.872} & 0.651  & 0.431  & 0.427  \\
        & 20    & 50    & 0.859  & 0.648  & 0.440  & \textbf{0.431} \\
        & 30    & 30    & 0.868  & 0.658  & 0.446  & 0.430  \\
        & 30    & 50    & 0.868  & 0.659  & \textbf{0.450} & \textbf{0.431} \\
        & 40    & 30    & 0.868  & \textbf{0.666} & 0.428  & 0.423  \\
        & 40    & 50    & 0.868  & \textbf{0.666} & 0.431  & 0.423  \\
        \midrule
        \multirow{10}[2]{*}{AVIN-TF} & 10    & 30    & 0.562  & 0.506  & 0.413  & 0.420  \\
        & 10    & 50    & 0.546  & 0.502  & 0.419  & 0.422  \\
        & 20    & 10    & 0.791  & 0.607  & 0.463  & 0.441  \\
        & 20    & 30    & 0.791  & 0.598  & \textbf{0.478} & \textbf{0.449} \\
        & 20    & 50    & 0.795  & 0.604  & \textbf{0.478} & 0.447  \\
        & 30    & 30    & \textbf{0.843} & 0.632  & 0.448  & 0.433  \\
        & 30    & 50    & 0.827  & 0.634  & 0.446  & 0.433  \\
        & 40    & 30    & 0.807  & 0.639  & 0.409  & 0.413  \\
        & 40    & 50    & 0.819  & \textbf{0.643} & 0.392  & 0.406  \\
        & -     & -     & 0.755  & 0.594  & 0.459  & 0.441  \\
        \bottomrule
    \end{tabular}
    \label{tab_abl_tptn}
\end{table}

\section{Conclusion}
In this work, Induction Network is proposed to bridge the gap between audio and visual modalities. After decoupling the gradients of different modalities, the audio and visual representations are aligned by the Induction Vector, which is obtained from the visual modality in a bootstrap manner. Nevertheless, an adaptive threshold selection strategy and a visually weighted contrastive loss are proposed to further improve the robustness of the network.

\textbf{Limitations}. Although visual weighted contrastive loss is used to correct faulty negatives, the faulty positives in the training set, i.e., audio-visual irrelevant pairs, also limit the localization performance. A faulty positive mining approach is considered to mitigate this issue.

\begin{acks}
This work is supported by the National Natural Science Foundation of China (No. 61971352, No. 62271239), Ningbo Natural Science Foundation (No. 2021J048, No. 2021J049), Jiangxi Double Thousand Plan (No. JXSQ2023201022), Fundamental Research Funds for the Central Universities (No. D5000220190), Innovative Research Foundation of Ship General Performance (No. 25522108).
\end{acks}

\bibliographystyle{ACM-Reference-Format}
\balance
\bibliography{references}

\newpage
\appendix

\section{Appendix}

\subsection{Datasets}
\textbf{SoundNet-Flickr}: SoundNet-Flickr \cite{thomee2016yfcc100m} is a dataset consisting of over 2 million real-life image-sound pairs with 500 annotated bounding boxes by Senocak et al. \cite{senocak2019learning}, and each pair is processed by 3 annotators. Following \cite{song2022self}, the training set contains a random subset of 10k and 144k pairs, while the testing set contains 250 annotated pairs.

\noindent\textbf{VGG-Sound and VGG-Sound Source}: VGG-Sound \cite{chen2020vggsound} is a more challenging dataset that includes over 200k in-the-wild video clips from YouTube with 10s audio and video segments. VGG-Sound Source (VGG-SS) is an audio-visual localization benchmark with 5k bounding box annotations of manually verified sounding objects \cite{chen2021localizing}. Following \cite{song2022self}, the training set for both datasets includes a random subset of 10k and 144k pairs, while the testing set contains a subset (4664 samples) of VGG-SS due to some unavailable videos.

\subsection{Implementation Details}
\textbf{Visual Network}: The ResNet18 \cite{he2016deep} and ViT-S \cite{dosovitskiy2020image} pretrained on ImageNet \cite{deng2009imagenet} are employed as visual encoder. ResNet18 comprises 8 residual blocks and the output channel is 512, ViT-S contains 12 blocks with the 6 heads and 384 channel in MHSA. The length of the image patch is 16. The parameters of ResNet18 are updated while those of ViT-S remain frozen during training. The visual projector projects the output channel to 512 for both ResNet and Transformer .

\noindent\textbf{Audio Network}: The ResNet22 model pretrained by PANNs \cite{kong2020panns} is employed as the audio encoder, including 8 residual blocks and the channel number of output embedding is 2048. The parameters of the audio encoder are frozen during training, and the output of the audio projector is a 512D vector.

\noindent\textbf{Data}: For data requirement, the middle frame of video clip and the 3-second sound surrounding the frame are selected as the visual and audio input, respectively. We performed operations of $224\times224$ random cropping and random horizontal flipping on the input images for data augmentation during training. Following \cite{kong2020panns}, the audio signal is resampled to 32kHz, and STFT is applied on waveforms with a Hamming window size of 1024 and a hop size of 320. The log-mel spectrogram is computed by applying 64 mel-filter banks, which is transformed to $301\times64$.

\noindent\textbf{Training Details}: The Adam optimizer is used with the rate of $1\times{10}^{-5}$ for the ResNet18 backbone, as well as the learning rate of $1\times{10}^{-4}$ for both the visual and audio projectors. The a batch size is set to 256 for training, and early stopping is configured to avoid overfitting. The hyperparameter of adaptive threshold selection strategy is set to $t_p=30$ and $t_n=50$ for AVIN-RN, $t_p=30$ and $t_n=30$ for AVIN-TF.

\subsection{Evaluation Metrics}
Following \cite{senocak2019learning}\cite{chen2021localizing}\cite{song2022self}, the metrics of cIoU and AUC are used for performance evaluation. The score map $g$ is computed for each sample, which is defined as:
\begin{equation}
g=\min\left(\sum_{j=1}^{n}\frac{b_j}{C},1\right)\label{eq19}
\end{equation}
where $b_j$ is the binary image of $j$-th bounding box, $C$ is the minimum number of opinions to reach an agreement and  $C=2$ in practice, thus cIoU is defined as:
\begin{equation}
cIoU\left(t\right)=\frac{\sum_{i\in A\left(t\right)} g_i}{\sum_{i} g_i+\sum_{i\in A\left(t\right)-G}1}\label{eq20}
\end{equation}
where $i$ is the pixel index of the score map, and the decision threshold $t$ is set to $0.5$. $A\left(t\right)=\{i|s_i>t\}$ and $G=\{i|g_i>0\}$, where $s_i$ indicates the activation of heatmap $S$ at location $i$. AUC is the area under the curve plotted by the ratio of samples with $cIoU>t{'}$ to the total number of samples when $t{'}$ changing from 0 to 1.

\subsection{Computational Complexity Analysis}
We perform the computational complexity analysis for different methods as shown in Table~\ref{tab_cpx_als}. The number of parameters (column 5) for AVIN-RN is 76.4M, whereas AVIN-TF is 89.5M (where ResNet22 \cite{kong2020panns} contributes 63.6M). It is worth noting that the SSPL has more parameters than the proposed model (108.6M vs. 76.4M/89.5M) but exhibits lower performance.

Additionally, the floating point operations (FLOPs) for each method (column 4) is also calculated, and the AVIN is able to maintain a modest computational cost as well as to achieve a better performance simultaneously.

Finally, we compare the speed of training (column 2) and inference (column 3) for all the methods. The training process is running on two 2080Ti GPUs, whereas inference is on a single 2080Ti GPU. It is normal for certain models to exhibit faster training speed compared to inference, which is owing to the acceleration of two GPUs.

\begin{table*}
  \centering
  \caption{Computational complexity analysis of different methods}
    \begin{tabular}{ccccc}
    \toprule
    Method & Train (AVpairs/s) & Inference (AVpairs/s) & FLOPs & Param \\
    \midrule
    HardWay \cite{chen2021localizing} & 474   & 430   & 5.5G  & 23.4M \\
    SSPL \cite{song2022self} & 90    & 225   & 46.7G & 108.6M \\
    FNAC \cite{sun2023learning} & 492   & 394   & 5.6G  & 22.9M \\
    FNAC+OGL \cite{sun2023learning} & 492   & 351   & 7.4G  & 34M \\
    AVIN-RN & 580   & 560   & 10.6G & 76.4M \\
    AVIN-TF & 156   & 385   & 12.3G & 89.5M \\
    \bottomrule
    \end{tabular}%
  \label{tab_cpx_als}%
\end{table*}%

Among the proposed models, AVIN-RN demonstrates the fastest training speed, which can process approximately 580 audio-visual pairs per second. AVIN-TF exhibits a relatively slower training speed, approximately 156 pairs per second, but still faster than SSPL. This discrepancy is affected by the TokenCut \cite{wang2022tokencut} that runs on the CPU and lacks acceleration, and thus consequently requires a longer duration. In the inference stage, AVIN-RN maintains its superiority in terms of speed, and achieves around 560 audio-visual pairs per second. Comparatively,  AVIN-TF achieves 385 pairs per second, which is similar as FNAC and faster than SSPL. 

\subsection{Visualization results}
\label{vis}
\begin{figure*}
  \includegraphics[width=\textwidth]{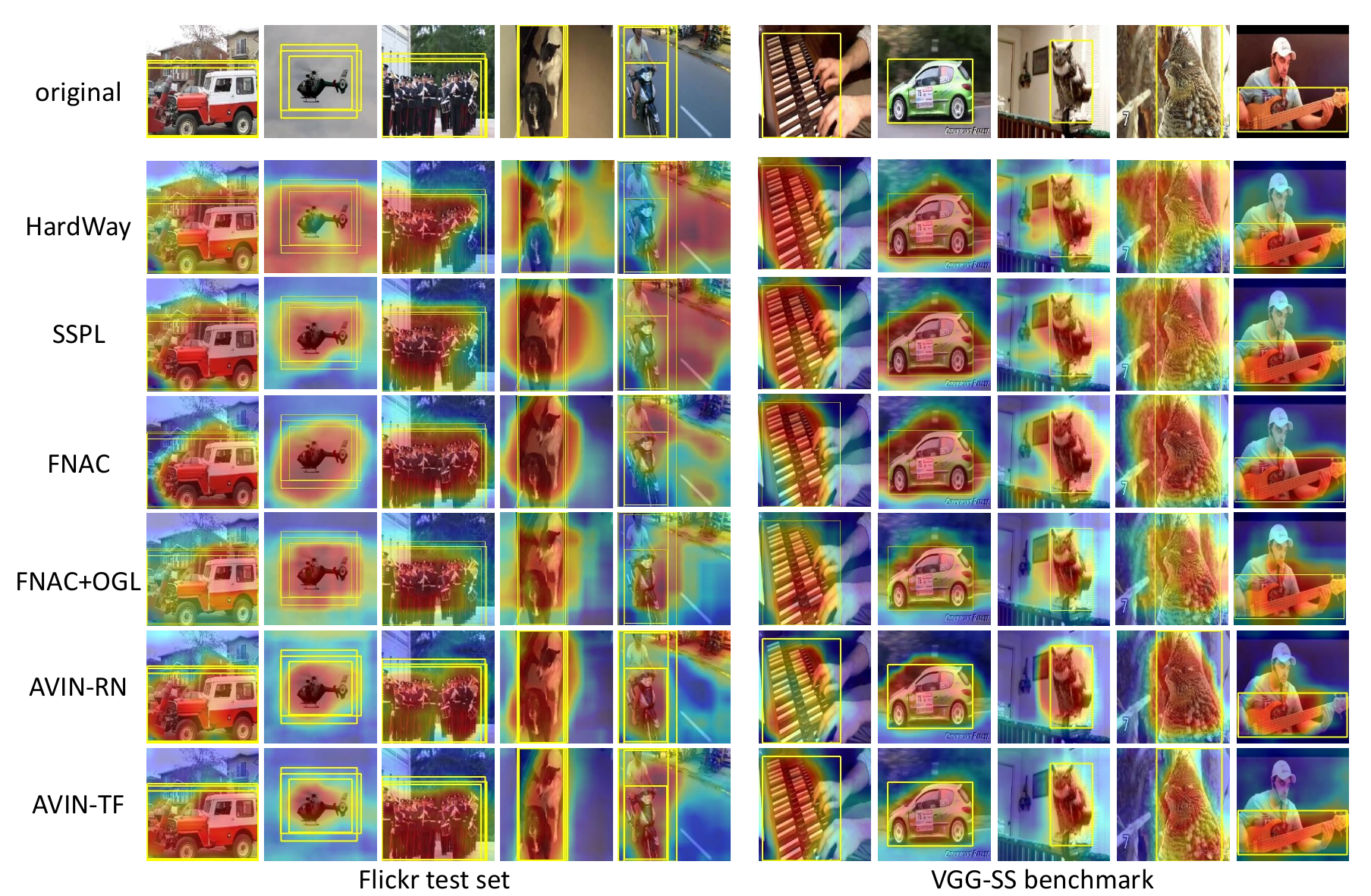}
  \caption{Visualization of different methods on Flickr test set and VGG-SS benchmark. Both AVIN-RN and AVIN-TF are able to localize sound sources in a variety of challenging scenarios.}
  \label{fig_visualization}
\end{figure*}

In this section, visualization results of HardWay \cite{chen2021localizing}, SSPL \cite{song2022self}, FNAC \cite{sun2023learning}, FNAC+OGL, AVIN-RN and AVIN-TF are shown in Figure~\ref{fig_visualization}.

\end{document}